\documentclass[11pt]{article}

\usepackage{cuted}
\usepackage{caption}
\usepackage{tcolorbox}
\tcbuselibrary{breakable, skins}
\newcounter{casestudy}

\usepackage[preprint]{acl}

\usepackage{times}
\usepackage{latexsym}
\usepackage{pifont}

\usepackage[T1]{fontenc}

\usepackage[utf8]{inputenc}

\usepackage{microtype}

\usepackage{inconsolata}
\usepackage{caption}
\usepackage{graphicx}
\usepackage{paralist}
\usepackage{tcolorbox}
\tcbuselibrary{breakable, skins}
\usepackage{arydshln}
\usepackage{enumitem}
\usepackage{booktabs}
\usepackage{multirow}
\usepackage[table]{xcolor}

\definecolor{mygray}{gray}{0.9} 
\definecolor{myblue}{HTML}{ECF4FF}
\usepackage{amsmath}
\usepackage{tcolorbox}
\usepackage{stfloats}
\usepackage{algorithm}
\usepackage{algorithmic}

\title{HybridThinker: Efficient Chain-of-Thought Reasoning via Compressed Memory and Transient Thought Steps}

\author{
  \textbf{Xin Liu\textsuperscript{1}\thanks{Equal contribution.}},
  \textbf{Runsong Zhao\textsuperscript{1}\footnotemark[1]},
  \textbf{Xinyu Liu\textsuperscript{1}},
  \textbf{Junhao Ruan\textsuperscript{1}},
  \textbf{Pengcheng Huang\textsuperscript{1}},
  \textbf{Shichao Dong\textsuperscript{1}},
  \\
  \textbf{Chunyang Xiao},
  \textbf{Chenglong Wang\textsuperscript{1}},
  \textbf{Changliang Li\textsuperscript{2}},
  \textbf{Jingbo Zhu\textsuperscript{1,3}},
  \textbf{Tong Xiao\textsuperscript{1,3}}\thanks{Corresponding author.}
  \\
  \textsuperscript{1} \normalsize{School of Computer Science and Engineering, Northeastern University, Shenyang 110819, China} \\
  \textsuperscript{2} \normalsize{China Unicom Cloud-Link, China} \\
  \textsuperscript{3} \normalsize{NiuTrans Research, Shenyang, China} \\
  \normalsize{\{liuxin1, zhaors\}@mails.neu.edu.com} \\
  \normalsize{\{xiaotong, zhujingbo\}@mail.neu.edu.com}
}

\begin{document}
\maketitle

\begin{abstract}

Extended chain-of-thought (CoT) traces improve LLM reasoning but incur substantial computational and memory costs.
While existing CoT compression methods mitigate this by condensing thought steps into compact representations via memory tokens and retaining only these representations at inference time, the loss of fine-grained information makes subsequent steps more error-prone. 
To alleviate this, we propose \textbf{HybridThinker}, where in addition to preserved these representations, thought steps are also temporarily retained to provide fine-grained details.
However, we observe that naively keeping thought steps accessible to subsequent steps \emph{during training} lets the model bypass memory tokens by retrieving information directly from these steps, leaving the model's ability to compress and retrieve information through memory tokens insufficiently trained.
We therefore introduce a hybrid training scheme, in which only some thought steps are directly accessible through attention to subsequent steps, while the other thought steps are masked, forcing the model to use memory tokens for compression and retrieval.
Across 4 reasoning benchmarks, HybridThinker matches the uncompressed baseline, advancing the state of the art in CoT compression by 5.8 points on average accuracy with similar inference time. Ablation studies confirm that both temporary thought-step retention and the hybrid training scheme contribute to these gains.

\end{abstract}

\section{Introduction}

\begin{figure*}[t]
    \centering
    \includegraphics[width=1.0\linewidth]{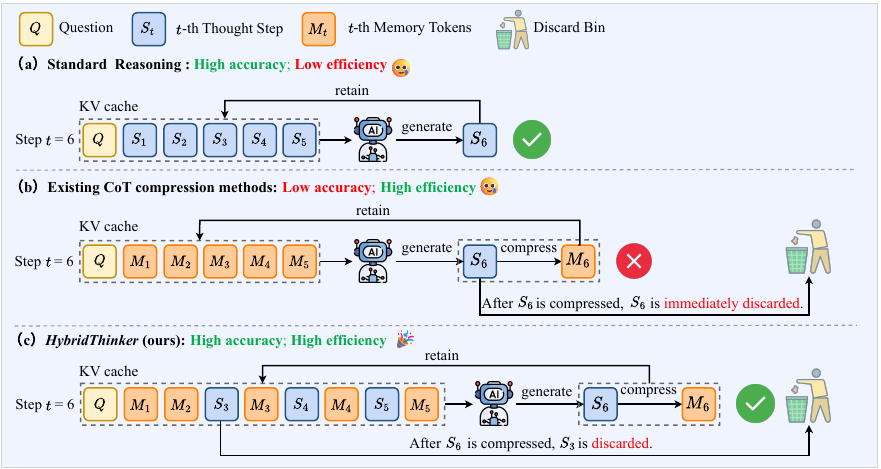}
    \caption{Comparison of three reasoning paradigms in terms of KV cache management. (a) Standard reasoning retains the KV cache of all original thought steps throughout decoding, achieving high accuracy at the cost of efficiency. (b) Existing CoT compression methods discard the KV cache of original thought steps immediately after compression, reducing memory usage but losing fine-grained local information. (c) HybridThinker (ours) temporarily retains the KV cache of each original thought step to serve several subsequent steps before discarding it, achieving both high accuracy and high efficiency. In this example, the KV cache of the third thought step $S_3$ is retained to serve several subsequent steps until $S_6$ has been compressed.}
    \label{fig:main}
\vspace{-5pt}
\end{figure*}

Recent advances in large language model reasoning have increasingly relied on extended chain of thoughts (CoTs), which have been shown to be effective across arithmetic, common sense, and knowledge-intensive tasks~\citep{wei2023chainofthoughtpromptingelicitsreasoning,kojima2022large,yao2023tree,Guo_2025,qwen2024qwen25}. However, as illustrated in Figure~\ref{fig:main}(a), these extended CoT traces introduce substantial inference overhead, as the self-attention computation scales quadratically and the KV cache grows linearly with the sequence length, hindering the broader deployment of reasoning models.

To alleviate this, two main approaches have been proposed: KV cache pruning, and CoT compression. KV cache pruning reduces memory~\cite{zhang2023h2o, chen2025sepllmacceleratelargelanguage} usage by selectively retaining important portions of the KV cache while discarding less critical ones during inference, but discarded information is irrecoverable without full regeneration. 
CoT compression~\citep{zhang2025lightthinker,kontonis2026mementoteachingllmsmanage,liu2026memosightunifyingcontextcompression, zhu2026lightthinkerreasoningcompressionmemory} condenses lengthy thought steps into compact representations, after which the original thought steps are discarded immediately as we illustrate in Figure~\ref{fig:main}(b). Unfortunately, this premature discarding renders the immediately following step more error-prone: key details such as exact numerical values, strict constraints, and intermediate conclusions may be obscured or lost in the compact representations~\citep{zhang2025latent,hybridcot2026}. In multi-step reasoning scenarios, such single-step errors propagate along the chain of thought in a cascading fashion. 

To enable efficient reasoning in LLMs, we propose \textbf{HybridThinker}, a CoT compression method that mitigates the loss of fine-grained local information caused by the immediate discarding of thought steps in existing CoT compression methods. Similar to existing CoT compression approaches, HybridThinker compresses lengthy thought steps into compact representations via memory tokens and retains those representations after each thought step is completed. However, the thought steps are not immediately discarded in HybridThinker; instead, they are temporarily retained and remain accessible to several subsequent steps. We illustrate the HybridThinker inference in Figure~\ref{fig:main}(c). 

To train HybridThinker, we observe that naively applying the same attention pattern during training as at inference is suboptimal (see Section~\ref{section:ablation_studies} for details) due to shortcut learning: when both memory tokens and thought steps are visible to subsequent steps, the model over-relies on the more readable plain-text thoughts, leaving its ability to compress and retrieve information through memory tokens undertrained.
We therefore propose \textbf{Hybrid Attention} for training, combining two complementary mechanisms: \textbf{Bottleneck Attention}, which hides each thought step from subsequent steps, forcing the model to access prior reasoning exclusively through memory tokens; and \textbf{Shortcut Attention}, which keeps each thought step visible to the immediately following few steps before discarding it, adapting the model to the inference-time setting. 

Experiments demonstrate that HybridThinker achieves an average accuracy improvement of 5.8 points over existing CoT compression methods, matching the performance of the uncompressed baseline.
Ablation studies further confirm the effectiveness of Hybrid Attention as a training strategy, showing that it outperforms training with pure Bottleneck Attention or pure Shortcut Attention alone.
To summarize our contributions:
\begin{inparaenum}[(1)]
\item We propose HybridThinker, in which thought steps are temporarily retained and kept visible to several subsequent steps alongside compact representations during inference, alleviating the loss of fine-grained information.
\item We propose Hybrid Attention for training, which resolves the shortcut learning problem that arises in naive HybridThinker training.
\item We demonstrate that HybridThinker achieves substantial performance improvements over existing CoT compression methods, establishing a new state-of-the-art performance in CoT compression.
\end{inparaenum}

\section{Related Work}


We organize related work into two categories: KV cache pruning, which reduces memory by directly discarding less important KV cache entries, and compression techniques, which also discards the original KV cache but first compresses the discarded information into compact representations. Compression techniques can be further divided into prompt compression, a well-established research area, and CoT compression, an emerging direction.

\paragraph{KV Cache Pruning.}
KV cache pruning reduces memory footprint by discarding less important KV cache entries, with methods differing primarily in how importance is estimated. H$_2$O~\citep{zhang2023h2o} uses accumulated attention weights as the importance criterion. StreamingLLM~\citep{xiao2024streamingllm} treats the first token and the most recent tokens as important due to the attention sink phenomenon. SepLLM~\citep{chen2025sepllmacceleratelargelanguage} extends StreamingLLM by additionally retaining delimiter tokens, which implicitly compress contextual information. However, these methods face inherent limitations: low cache budgets have been shown to paradoxically lengthen CoT outputs~\citep{liu2025holdthoughtassessingkv}, and discarded cache becomes inaccessible.

\paragraph{Compression Techniques.}

Existing prompt compression methods~\citep{chang2024efficientpromptingmethodslarge,li2024promptcompressionlargelanguage,jiang2023llmlinguacompressingpromptsaccelerated, mu2024learningcompresspromptsgist, ge2024incontextautoencodercontextcompression, li2024500xcompressorgeneralizedpromptcompression, wang2024incontextformerlightningfastcompressing, tang2025perceptioncompressortrainingfreeprompt, 
tang2026readhumancompressingcontext, tang2026comicoarsetofinecontextcompression, tang2026gmsaenhancingcontextcompression, zhao-etal-2025-position, zhao2026cometcollaborativememorytransformer, zhao2026pocperformanceorientedcontextcompression, liu2026autoencodingfreecontextcompressionllms} focus on compressing model 
\emph{inputs}, which are fully accessible ahead of time. CoT reasoning, however, is generated incrementally and cannot be accessed in its entirety beforehand, necessitating step-wise compression and rendering most prompt compression techniques directly inapplicable. To fill this gap, several dedicated CoT compression methods have been proposed. LightThinker~\citep{zhang2025lightthinker} compresses each completed thought step into the KV cache of a small set of memory tokens and immediately discards the thought step's KV cache, retaining only the memory tokens. MEMENTO~\citep{kontonis2026mementoteachingllmsmanage} instead compresses thought steps into shorter text sequences. MemoSight~\citep{liu2026memosightunifyingcontextcompression} improves upon LightThinker's compression mechanism and further incorporates multi-token prediction. However, all of these CoT compression methods discard thought steps immediately upon compression, losing fine-grained local information.

\section{Method}

In this section, we introduce HybridThinker.
We first define the key notation in Section~\ref{sec:notation}.
Then, in Section~\ref{sec:inference}, we describe the inference procedure of HybridThinker from a KV cache management perspective.
In Section~\ref{sec:training}, we present the training scheme of HybridThinker, which addresses the shortcut learning problem from directly applying the inference‑time attention pattern during training.

\subsection{Notation}
\label{sec:notation}

We introduce the key notation used in Section~\ref{sec:inference} and Section~\ref{sec:training}. We use $t = 1, 2, \ldots, k$ to index thought steps. $w$ is the temporary retention duration of a thought step. For example, $w=4$ in Figure~\ref{fig:main}(c) as $S_3$ is accessible to $S_3, S_4, S_5, S_6$. 


\paragraph{Input and Output.}
The model receives a request \(Q\) that requires CoT reasoning and generates a CoT \(Y = S_1, S_2, \ldots, S_k\) consisting of \(k\) thought steps. 
The answer is contained in \(Y\).
Each thought step $S_t$ is a token sequence terminated by a delimiter.\footnote{Following \citet{zhang2025lightthinker}, we use \texttt{``\textbackslash n\textbackslash n''} as the step delimiter in this work.}

\paragraph{Memory Tokens.}
HybridThinker maintains a fixed sequence of $L$ learnable special tokens
$M = (m_1, \ldots, m_L)$,
called \emph{memory tokens}.
After generating $S_t$, the model performs a forward pass over $M$ to produce the KV cache $\tilde{M}_t$, which serves as a compressed summary of $S_t$.
A fresh copy of $M$ is used at each thought step to capture the thought step semantics; we write $M_t$ when we specifically refer to the memory tokens inserted after $S_t$.

\paragraph{KV Cache Container.}
We use $\mathcal{C}_t$ to denote the KV cache container 
that the model attends to when generating step $S_t$.
We introduce three types of KV cache notation used throughout:
\begin{inparaenum}[(1)]
    \item $\tilde{Q}$: the KV cache of the question $Q$, computed once during prefill and retained throughout.
    \item $\tilde{M}_t$: the KV cache of memory tokens $M_t$ produced by compressing $S_t$.
    \item $\tilde{S}_t$: the KV cache of the raw thought step $S_t$ that is retained temporarily through a few subsequent thought steps in HybridThinker.
\end{inparaenum}

\begin{figure*}[t]
    \centering
    \includegraphics[width=1.0\linewidth]{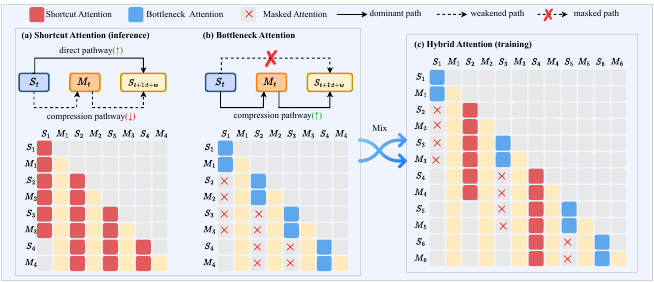}
    \caption{Three attention masks in HybridThinker, where $S_t$ and $M_t$ denote the $t$-th thought step and the memory tokens responsible for compressing it, respectively; $S_{t:t'}$ denotes $S_t,S_{t+1}, ..., S_{t'-1}$. (a) \textbf{Shortcut Attention} (HybridThinker inference): (bottom) $S_t$ is visible to the following $w-1$ steps (here $w=3$), but during training (top) this direct pathway becomes dominant and overshadows compression pathway through $M_t$. (b) \textbf{Bottleneck Attention}: the direct pathway is masked, forcing subsequent steps to retrieve information exclusively through $M_t$. (c) \textbf{Hybrid Attention} (HybridThinker training): Shortcut Attention is applied to a random subset of steps (here $S_2$ and $S_4$) and Bottleneck Attention to the rest, training both pathways simultaneously. The question $Q$ is omitted.}
    \label{fig:attention_mask}
\vspace{-5pt}
\end{figure*}

\subsection{HybridThinker Inference}
\label{sec:inference}
 
HybridThinker generates CoT $Y$ by repeating the following three steps for each thought step indexed by $t = 1, 2, \ldots, k$.
 
\paragraph{Step 1: Generate $S_t$.}
The LLM parameterized by $\theta$ generates thought step $S_t$ autoregressively, conditioning on the KV cache container $\mathcal{C}_t$:
\begin{equation}
    S_t \sim P_\theta(\,\cdot \mid \mathcal{C}_t).
\end{equation}
Initially, \(\mathcal{C}_1\) is set to \(\{\tilde{Q}\}\), i.e., the KV cache of the question $Q$. As a by-product of autoregressive decoding, the KV cache $\tilde{S}_t$ of $S_t$ is stored.
 
\paragraph{Step 2: Compress $S_t$.}
Once $S_t$ is complete, the model compresses its content via memory tokens $M_t$ by performing a forward pass:
\begin{equation}
    \tilde{M}_t = \mathrm{LLM}_\theta(M_t \mid \mathcal{C}_t,\, \tilde{S}_t),
\end{equation}
where the model attends to both the currently retained KV cache $\mathcal{C}_t$ and the newly generated $\tilde{S}_t$ to produce the compressed KV cache $\tilde{M}_t$.
Intuitively, $\tilde{M}_t$ encodes the information in $S_t$ into a compact representation that can be accessed by future steps.
 
\paragraph{Step 3: Update $\mathcal{C}_t$.}
After compression, we update the KV cache container for the next step.
The question cache $\tilde{Q}$ and all memory token caches $\tilde{M}_1, \ldots, \tilde{M}_t$ are always retained, as in existing CoT compression methods.
Unlike existing methods, HybridThinker also temporarily retains 
$\tilde{S}_t$, serving the next $w-1$ steps with fine-grained 
information:
\begin{equation}
\mathcal{C}_{t+1} =
\begin{cases}
\mathcal{C}_t \cup \{\tilde{S}_t,\, \tilde{M}_t\} & \text{if } 1 \leq t < w \\[8pt]
\begin{aligned}
& \bigl(\mathcal{C}_t \setminus \{\tilde{S}_{t-w+1}\}\bigr) \\
& \qquad \cup \;\{\tilde{S}_t,\, \tilde{M}_t\}
\end{aligned} & \text{if } t \geq w
\end{cases}
\end{equation}
where $\setminus$ denotes set difference.
In other words, in general (i.e. for $t \geq w$), once a thought step is completed with its associated memory tokens generated: 1) $\tilde{S}_t$ and $\tilde{M}_t$ are pushed into the KV cache container $C_t$ for the subsequent CoT generations 2) the thought step KV cache $\tilde{S}_{t-w+1}$ that was generated $w-1$ steps before is popped out from the container $C_t$.
Figure~\ref{fig:main}(c) illustrates the inference procedure of HybridThinker.

\subsection{HybridThinker Training}
\label{sec:training}

We fine-tune an LLM with CoT data to obtain a model capable of the inference procedure above.
Following other CoT compression methods, our approach requires only two modifications over standard LLM training: reconstructing the training data and adjusting the attention mask.
 
\paragraph{Data Reconstruction.}
Each training instance is a pair $(Q, Y)$.
Following \citet{zhang2025lightthinker}, we segment $Y$ into $k$ thought steps $S_1, \ldots, S_k$ using \texttt{``\textbackslash n\textbackslash n''} as the delimiter, then insert a memory token sequence $M_t$ after each intermediate step $S_t$ ($t < k$) to form the training sequence:
\begin{equation}
\begin{split}
X &= Q \oplus S_1 \oplus M_1 \oplus S_2 \oplus M_2 \oplus \cdots \oplus S_k \\
  &= (x_1, x_2, \ldots, x_{|X|}),
\end{split}
\end{equation}
where $\oplus$ denotes concatenation and $x_u$ is the $u$-th token of $X$.
All $M_1, \ldots, M_{k-1}$ share the same token content but occupy different positions.

\begin{table*}[!t]
\centering
\tiny
\setlength{\tabcolsep}{1.5pt} 
\resizebox{\linewidth}{!}{

    \begin{tabular}{l ccc | ccc | ccc | ccc | ccc}
    \toprule
    
    \multirow{4}{*}{\textbf{Method}} & \multicolumn{3}{c}{\textbf{GSM8K}} & \multicolumn{3}{c}{\textbf{MMLU}} & \multicolumn{3}{c}{\textbf{GPQA}} & \multicolumn{3}{c}{\textbf{BBH}} & \multicolumn{3}{c}{\textbf{AVG.}} \\ 
    \cmidrule(lr){2-4} \cmidrule(lr){5-7} \cmidrule(lr){8-10} \cmidrule(lr){11-13} \cmidrule(lr){14-16}
    
    & Acc~$\uparrow$ & Time~$\downarrow$ & Peak~$\downarrow$ 
    & Acc~$\uparrow$ & Time~$\downarrow$ & Peak~$\downarrow$ 
    & Acc~$\uparrow$ & Time~$\downarrow$ & Peak~$\downarrow$ 
    & Acc~$\uparrow$ & Time~$\downarrow$ & Peak~$\downarrow$ 
    & Acc~$\uparrow$ & Time~$\downarrow$ & Peak~$\downarrow$ \\ 
    \cmidrule{1-16}

    \rowcolor{mygray} \multicolumn{16}{c}{\textit{Qwen2.5-7B Series}} \\ 
    \cmidrule{1-16}
    CoT &{88.32}&{3.49}&{519} &{70.01}&{3.88}&{676} &{28.28}&{1.18}&{998} &{71.31}&{1.27}&{582} &{64.48}&{2.46}&{694} \\
    Distill-R1 &{60.58}&{6.54}&{522} &{31.26}&{15.38}&{1289} &{22.22}&{11.89}&{4383} &{51.31}&{3.71}&{905} &{41.34}&{9.38}&{1775} \\ 
    \cmidrule{1-16}
    Vanilla &{90.83}&{22.19}&{1559} &{65.92}&{28.89}&{2241} &{37.37}&{24.50}&{7184} &{81.01}&{10.59}&{2212} &{68.78}&{21.54}&{3299} \\
    \rowcolor{myblue} 
    \quad + H$_2$O &{\underline{91.05}}&{37.98}&{1024} &\underline{62.32}&{76.01}&{1024} &{20.20}&{47.98}&\textbf{1024} &\textbf{}{75.76}&{25.01}&\underline{1024} &{62.33}&{46.75}&\textbf{1024} \\ 
    \rowcolor{myblue} 
    \quad + SepLLM &{\textbf{91.28}}&{49.03}&{1024} &{59.20}&{66.76}&{1024} &{12.63}&{40.89}&\textbf{1024} &{69.70}&{28.36}&\underline{1024} &{58.20}&{46.26}&\textbf{1024} \\ 
    \rowcolor{myblue} 
    LightThinker &{87.26}&\underline{23.05}&\textbf{684} &{60.66}&{\textbf{22.52}}&\textbf{827} &\underline{37.37}&{\textbf{13.33}}&\underline{1969} &{66.46}&\underline{10.91}&\textbf{980} &\underline{62.94}&{\underline{17.45}}&\underline{1115} \\ 
    \cmidrule{1-16}
    \rowcolor{myblue} 
    \textbf{HybridThinker} &{90.60}&{\textbf{20.85}}&\underline{864} &{\textbf{63.39}}&\underline{23.07}&\underline{1009} &\textbf{47.98}&\underline{14.84}&\underline{2124} &\underline{73.13}&{\textbf{9.87}}&{1058} &\textbf{68.78}&\textbf{17.15}&{1264} \\ 

    \cmidrule{1-16}
    \rowcolor{mygray} \multicolumn{16}{c}{\textit{Llama3.1-8B Series}} \\ 
    \cmidrule{1-16}
    CoT &{64.52}&{3.63}&{497} &{60.18}&{5.13}&{698} &{24.75}&{4.58}&{2230} &{53.13}&{1.39}&{596} &{50.65}&{3.68}&{1005} \\ 
    Distill-R1 &{57.62}&{7.20}&{524} &{20.93}&{23.64}&{1611} &{30.81}&{18.16}&{5640} &{29.49}&{5.15}&{1146} &{34.71}&{13.54}&{2230} \\  
    \cmidrule{1-16}
    Vanilla &{89.31}&{27.89}&{1702} &{71.08}&{42.86}&{2800} &{37.37}&{24.72}&{6824} &{78.18}&{13.81}&{2515} &{68.99}&{27.32}&{3460} \\ 
    \rowcolor{myblue} 
    \quad + H$_2$O &\textbf{89.31}&{48.08}&{1024} &\textbf{69.52}&{90.18}&\underline{1024} &{23.23}&{53.47}&\textbf{1024} &\textbf{81.62}&{23.81}&\underline{1024} &\underline{65.92}&{53.89}&\textbf{1024} \\ 
    \rowcolor{myblue} 
    \quad + SepLLM &\underline{88.32}&{27.92}&{1024} &\underline{64.85}&{49.03}&\underline{1024} &{15.66}&{23.69}&\textbf{1024} &{72.32}&{15.48}&\underline{1024} &{60.29}&{29.03}&\textbf{1024} \\ 
    \rowcolor{myblue} 
    LightThinker &{85.82}&\underline{25.87}&\textbf{664} &{60.86}&\underline{33.45}&\textbf{932} &\underline{37.37}&\underline{15.59}&\underline{1933} &{72.93}&\underline{13.58}&\textbf{999} &{64.25}&\underline{22.12}&\underline{1132} \\  
    \cmidrule{1-16}
    \rowcolor{myblue}
    \textbf{HybridThinker} &{87.34}&\textbf{25.50}&\underline{878} &{62.71}&\textbf{31.54}&{1053} &\textbf{37.88}&\textbf{14.55}&{1971} &\underline{80.00}&\textbf{11.73}&{1031} &\textbf{66.98}&\textbf{20.83}&{1233} \\

    \bottomrule
    \end{tabular}
}

\caption{Main results on the Qwen2.5-7B and Llama3.1-8B models. Memory-efficient methods are highlighted in blue, with \textbf{bold} and \underline{underlined} values indicating the best and second-best results among all the memory-efficient methods. Peak and Time denote the maximum token count stored in KV cache and the total inference time over all samples, respectively.}
\label{table:exp_main}
\vspace{-5pt}
\end{table*}

\paragraph{Attention Mask Construction.}
The most natural way to construct the training attention mask is to replicate the inference-time attention pattern directly.
At inference time, information in each thought step $S_t$ reaches subsequent steps via two pathways: (i) an \emph{compression pathway}, in which $S_t$ passes coarse-grained information through the memory tokens $M_t$; and (ii) a \emph{direct pathway}, in which $S_t$ is directly attended to by the subsequent $w-1$ steps.
Directly replicating this pattern during training yields what we call \textbf{Shortcut Attention}:
as illustrated on the top of Figure~\ref{fig:attention_mask}(a), because the direct pathway provides a much easier route for the model to retrieve information about $S_t$, training with Shortcut Attention leads to a \emph{shortcut learning} problem: the model over-relies on the direct pathway and the compression pathway receives insufficient training signal. We illustrate the Shortcut Attention at the bottom of Figure~\ref{fig:attention_mask}(a) and use it in our experiments as the natural baseline.
 
To remedy this, one could instead mask out the direct pathway entirely, forcing all information to pass through $M_t$ as the sole bottleneck.
We call this \textbf{Bottleneck Attention}. In Figure~\ref{fig:attention_mask}(b), we illustrate its motivation (top) as well as its attention implementations (bottom).
While Bottleneck Attention effectively trains the compression pathway, it provides no training signal for the direct pathway that is active at inference time, leading to a train--inference mismatch.
 
To address the limitations of both attention patterns, we propose \textbf{Hybrid Attention} (Figure~\ref{fig:attention_mask}(c)), which randomly assigns each thought step to either Shortcut Attention or Bottleneck Attention within the same training sequence.
This ensures both pathways receive adequate training.

Concretely, we randomly partition the $k$ thought steps into two disjoint sets $\mathcal{S}_\mathrm{shortcut} = \{S_t \mid t \in \mathcal{I}\}$ and $\mathcal{S}_\mathrm{bottleneck} = \{S_t \mid t \notin \mathcal{I}\}$, where $\mathcal{I} \subseteq \{1,\ldots,k\}$ is sampled uniformly at random.
The resulting attention mask $\mathbf{A} \in \{0,1\}^{|X| \times |X|}$ is defined as:
\begin{equation}
\mathbf{A}_{u,v} =
\begin{cases}
1 & \text{if } x_v \in Q,\ v \leq u \\[8pt]
1 & \text{if } x_v \in M_t,\ v \leq u \\[8pt]
1 & \text{if } \parbox[t]{0.5\textwidth}{$x_v \in S_t,\ S_t \in \mathcal{S}_{\text{bottleneck}},$\\$x_u \in S_t \cup M_t,\ v \leq u$} \\[22pt]
1 & \text{if } \parbox[t]{0.5\textwidth}{$x_v \in S_t,\ S_t \in \mathcal{S}_{\text{shortcut}},\ v \leq u,$\\$x_u \in \bigcup_{t'=t}^{t+w-1}(S_{t'} \cup M_{t'})$} \\[22pt]
0 & \text{otherwise.}
\end{cases}
\end{equation}
Rows 1--2 ensure that $Q$ and all memory tokens are globally visible.
Row 3 implements Bottleneck Attention: $S_t$ is only visible within its own step and to the immediately following $M_t$.
Row 4 implements Shortcut Attention: $S_t$ is additionally visible to the next $w-1$ steps and their memory tokens.

\paragraph{Training Objective.}
We minimise the standard cross-entropy loss over the thought step tokens:
\begin{equation}
    \mathcal{L}(\theta) = -\sum_{x_u \in S_t}
    \log P_\theta\!\left(x_u \;\middle|\; \mathcal{X}_u\right),
\end{equation}
where $\mathcal{X}_u = \{x_v \mid \mathbf{A}_{u,v} = 1,\, v < u\}$ is the set of tokens at positions strictly preceding $u$ and visible to position $u$.
Tokens in $Q$ and $M_t$ are excluded from the loss.

\section{Experiments}

\subsection{Experimental Setup}

\paragraph{Baselines.}
We evaluate HybridThinker on two backbone models, Qwen2.5-7B~\citep{hui2024qwen2} and Llama3.1-8B~\citep{dubey2024llama}, against six representative baselines organized into three groups by methodology. (1) CoT methods: the standard instruction model prompted with CoT, and the distilled reasoning model Distill-R1~\citep{Guo_2025}. 
(2) Trained reasoning models: Vanilla and LightThinker~\citep{zhang2025lightthinker}, both initialized from Distill-R1 and further instruction fine-tuned on Bespoke-Stratos-17k under the same training configuration as HybridThinker.
(3) Training-free acceleration techniques: H$_2$O~\citep{zhang2023h2o} and SepLLM~\citep{chen2025sepllmacceleratelargelanguage} are two KV cache pruning strategies applied directly on top of Vanilla. Full configuration details for all baselines are provided in Appendix~\ref{appendix:baseline_details}.

\paragraph{Evaluation Metrics and Datasets.}
We evaluate HybridThinker on four reasoning benchmarks: GSM8K~\citep{cobbe2021training}, MMLU~\citep{hendrycks2021measuring}, GPQA~\citep{rein2023gpqa}, and BBH~\citep{suzgun2022challenging}. These benchmarks span mathematical reasoning, knowledge-intensive question answering that demands extended reasoning chains, and multi-step complex reasoning, allowing us to test the performance of compressed reasoning under different reasoning demands. We report accuracy (\textbf{Acc}) for task performance, total inference time (\textbf{Time}), and peak token usage (\textbf{Peak}). Together, these three metrics capture whether a compression method can reduce deployment overhead while preserving accuracy.

\paragraph{Implementation Details.}
All trainable methods are trained for 5 epochs with a batch size of 64 and a maximum training sequence length of 4096 tokens. 
For HybridThinker, the number of memory tokens $L=9$ (same as LightThinker), the temporary retention duration $w=4$, and the number of Shortcut Attention steps $|\mathcal{I}|=2$.
The maximum output length during evaluation is set to 10240 tokens. Further training details are provided in Appendix~\ref{appendix:train_details}.

\subsection{Main Results}
Table~\ref{table:exp_main} shows our main results. On Qwen2.5-7B, HybridThinker matches the accuracy of the uncompressed baseline Vanilla (68.78\%) while reducing peak token usage by 61.7\% and inference time by 20.3\%. On Llama3.1-8B, HybridThinker (66.98\%) also approaches Vanilla (68.99\%). Compared with LightThinker, which discards thought steps immediately after compression, HybridThinker achieves a clear accuracy advantage across both models and all four benchmarks, outperforming it by 5.8 points on Qwen2.5-7B and 2.7 points on Llama3.1-8B on average, at the modest cost of retaining only about 100–150 more peak tokens, which translates to around 10\% more peak token usage. Notably, this accuracy gain does not come at the expense of inference speed: HybridThinker generates shorter CoT sequences than LightThinker (see Section~\ref{sec:output}), which offsets the overhead of the additional retained tokens and yields comparable or even slightly faster inference. 
HybridThinker also outperforms KV cache pruning methods including H$_2$O and SepLLM
with generally significant less total inference time. 
These results demonstrate that temporarily retaining thought steps to supply fine-grained local information to several subsequent steps effectively improves accuracy over existing CoT compression, establishing HybridThinker as a strong CoT compression approach.

\begin{table*}[!t]
\centering
\tiny
\setlength{\tabcolsep}{1.5pt} 
\resizebox{\linewidth}{!}{

    \begin{tabular}{l ccc | ccc | ccc | ccc | ccc}
    \toprule
    
    \multirow{4}{*}{\textbf{Training attention}} & \multicolumn{3}{c}{\textbf{GSM8K}} & \multicolumn{3}{c}{\textbf{MMLU}} & \multicolumn{3}{c}{\textbf{GPQA}} & \multicolumn{3}{c}{\textbf{BBH}} & \multicolumn{3}{c}{\textbf{AVG.}} \\ 
    \cmidrule(lr){2-4} \cmidrule(lr){5-7} \cmidrule(lr){8-10} \cmidrule(lr){11-13} \cmidrule(lr){14-16}
    
    & Acc~$\uparrow$ & Time~$\downarrow$ & Peak~$\downarrow$ 
    & Acc~$\uparrow$ & Time~$\downarrow$ & Peak~$\downarrow$ 
    & Acc~$\uparrow$ & Time~$\downarrow$ & Peak~$\downarrow$ 
    & Acc~$\uparrow$ & Time~$\downarrow$ & Peak~$\downarrow$ 
    & Acc~$\uparrow$ & Time~$\downarrow$ & Peak~$\downarrow$ \\ 
    \cmidrule{1-16}

    \rowcolor{mygray} \multicolumn{16}{l}{\textit{Shortcut attention inference (Temporarily retain thought steps during inference)}} \\ 
    \cmidrule{1-16}
    
    \rowcolor{myblue} 
    Hybrid (HybridThinker) &\textbf{90.60}&{\underline{20.85}}&{864} &{\textbf{63.39}}&{23.07}&{1009} &\textbf{47.98}&\underline{14.84}&{2124} &\textbf{73.13}&{\textbf{9.87}}&{1058} &\textbf{68.78}&\textbf{17.15}&{1264} \\

    Shortcut &\underline{90.52}&{29.79}&{971} &\underline{63.10}&{35.18}&{1360} &\underline{43.94}&{19.50}&{3420} &\underline{69.90}&{16.10}&{1595} &\underline{66.87}&{25.14}&{1836} \\

    Bottleneck &{80.52}&\textbf{20.78}&{880} &{52.97}&{24.29}&{1036} &{34.34}&{16.33}&{2269} &{62.83}&\underline{10.59}&{1109} &{57.67}&{18.00}&{1324} \\

    \cmidrule{1-16}

    \rowcolor{mygray} \multicolumn{16}{l}{\textit{Bottleneck attention inference (Immediately discard thought steps during inference)}} \\ 
    \cmidrule{1-16}
    Bottleneck (LightThinker) &{87.26}&{23.05}&\underline{684} &{60.66}&{\textbf{22.52}}&\underline{827} &{37.37}&{\textbf{13.33}}&\textbf{1969} &{66.46}&{10.91}&\textbf{980} &{62.94}&{\underline{17.45}}&\textbf{1115} \\ 
    
    Hybrid &{87.79}&{22.82}&\textbf{667} &{60.56}&\underline{23.02}&\textbf{828} &{35.86}&{15.07}&{2312} &{68.89}&{11.70}&\underline{1025} &{63.28}&{18.15}&\underline{1208} \\

    \bottomrule
    \end{tabular}
}
\caption{Ablation study on training attentions and inference strategies (Qwen2.5‑7B). The upper and lower sections correspond to temporarily retaining and immediately discarding thought steps during inference, respectively. Within each section, we ablate the training attention type across rows.}
\label{table:ablation}
\vspace{-5pt}
\end{table*}

\subsection{Ablation Results}
\label{section:ablation_studies}

 We analyze the effects of temporarily retaining thought steps and the hybrid training scheme and show our results in Table~\ref{table:ablation}.

\paragraph{Temporarily retaining thought steps during inference improves reasoning accuracy.} When thought steps are temporarily retained during inference (i.e. Shortcut Attention inference), the method naively trained with the Shortcut Attention achieves an average accuracy of 66.87\%; HybridThinker with a carefully designed training method trained with the Hybrid Attention improves the average accuracy further to 68.78\%. Comparatively, LightThinker which discards thought steps immediately during inference (i.e. Bottleneck Attention inference) achieves significant lower average accuracy at 62.94\%. This confirms that having each complete thought step serve only a few subsequent steps already yields a significant accuracy improvement.

\paragraph{Training with Hybrid Attention achieves the best accuracy--efficiency trade‑off.} With shortcut inference, shortcut‑only training reaches reasonable accuracy (66.87\%) but suffers from much higher inference time (25.14 vs.\ 17.15) and peak token usage (1836 vs.\ 1264).
We hypothesize that unrestricted direct access during training prevents the model from learning to rely on compressed memory, causing redundant reasoning and a longer chain (see Section~\ref{sec:case_study_analysis}).
Hybrid training avoids this shortcut learning by jointly training both the direct and compression pathways, resulting in the highest accuracy (68.78\%) together with the lowest latency and peak memory among temporary retention settings. Even under bottleneck inference, hybrid training performs slightly better than bottleneck‑only training (63.28\% vs.\ 62.94\%), indicating that the hybrid pattern does not harm compression ability.

Overall, the results show that temporarily retaining thought steps is critical for accuracy, and the hybrid training scheme effectively optimizes both information pathways for superior performance.

\section{Analysis}

\subsection{Hyperparameter Analysis}
\label{sec:hyper}
\begin{figure*}[t]
    \centering
    \includegraphics[width=1.0\linewidth]{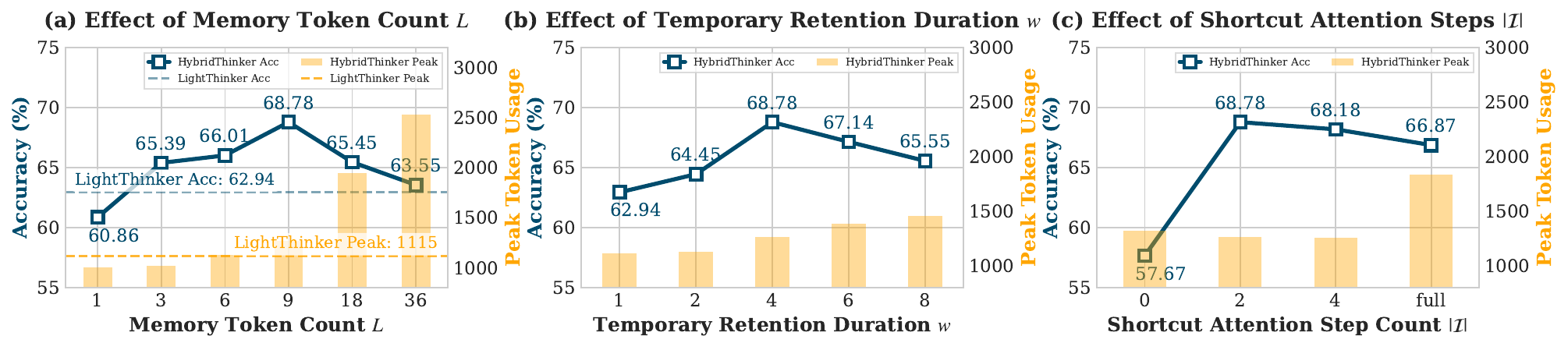}
    \caption{Hyperparameter sensitivity analysis on Qwen2.5-7B. Each subplot varies one hyperparameter while keeping the others at their defaults ($L=9$, $w=4$, $|\mathcal{I}|=2$): (a) memory token count $L$, (b) temporary retention duration $w$, and (c) number of Shortcut Attention steps $|\mathcal{I}|$.}
    \label{fig:hyperparameter_analysis}
\vspace{-5pt}
\end{figure*}

\paragraph{Effect of Memory Token Count $(L)$.} As shown in Figure~\ref{fig:hyperparameter_analysis}(a), the number of memory tokens determines the compression capacity for each thought step. Experiments show that average accuracy increases from $60.86\%$ to $68.78\%$ as $L$ grows from 1 to 9, reaching parity with the uncompressed baseline Vanilla at $L = 9$, which indicates that 9 memory tokens is sufficient to carry the information of one step. We remark that HybridThinker surpasses LightThinker accuracy with less peak token usage at $L=3$. Further increasing $L$ leads to accuracy degradation while peak token usage continues to rise, suggesting that additional memory tokens introduce redundant representations and context overhead rather than useful information.


\paragraph{Effect of Temporary Retention Duration $(w)$.} As shown in Figure~\ref{fig:hyperparameter_analysis}(b), starting from no thought step retention (\(w=1\)), a short retention duration (\(w=2\)) improves average accuracy but is still suboptimal to keep fine-grained information, yielding only 64.45\% accuracy. Extending the duration to the default setting (\(w=4\)) preserves sufficient fine-grained information and raises accuracy to 68.78\%. Further prolonging the retention duration makes such information arguably increasingly redundant, even noisy, causing accuracy to drop to 67.14\% and 65.55\%, while peak token usage keeps growing.


\paragraph{Effect of the Number of Shortcut Attention Steps (\(|\mathcal{I}|\)).} As shown in Figure~\ref{fig:hyperparameter_analysis}(c), $|\mathcal{I}|$ controls how many thought steps per training instance follow the Shortcut Attention pattern.
When $|\mathcal{I}|=0$ (pure Bottleneck Attention training), average accuracy drops to 57.67\%, reflecting the mismatch with the shortcut inference pattern.
Increasing $|\mathcal{I}|$ from 2 to 4 results in a slight decrease in average accuracy (68.78\% vs.\ 68.18\%), suggesting even a few shortcut steps suffice for the model to adapt to the Shortcut Attention pattern at inference time.
Training with only Shortcut Attention ($|\mathcal{I}| = \text{full}$) leads to worse average accuracy of 66.87\%, and also incurs significantly higher peak token usage. This is because the memory tokens are under-trained for compression and retrieval, causing the model to repeatedly re-read the question instead of continuing reasoning, as detailed in Section~\ref{sec:case_study_analysis}.

\subsection{Output Length Analysis}
\label{sec:output}
\begin{figure}[h]
    \centering
    \includegraphics[width=1.0\linewidth]{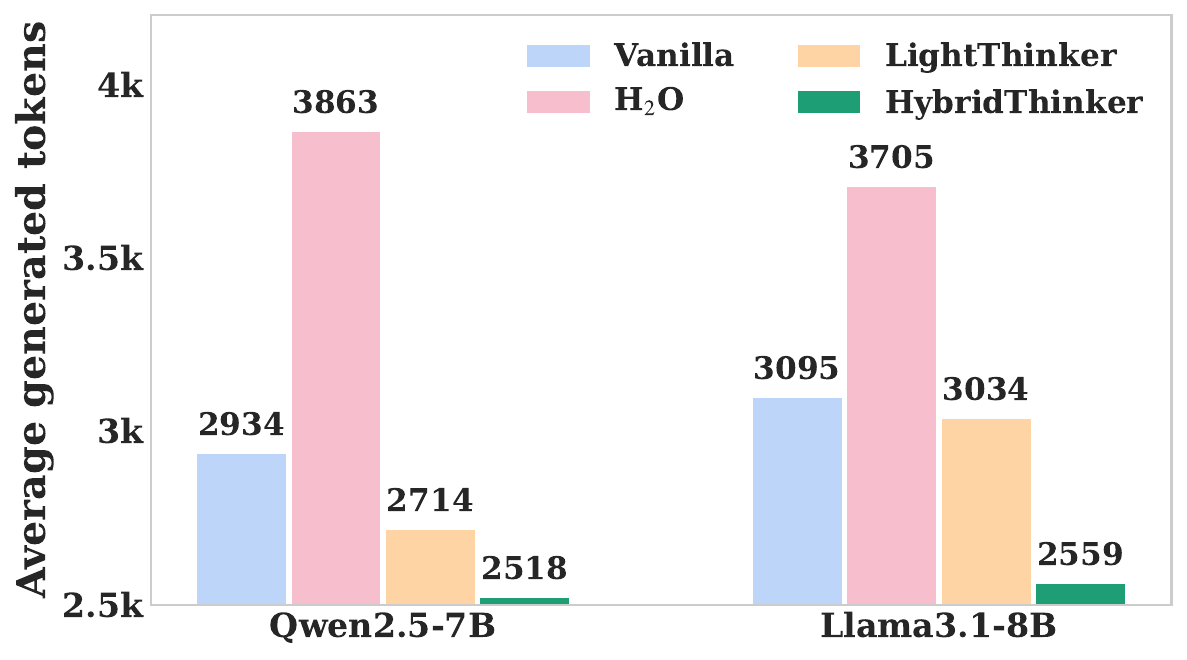}
    \caption{Average number of generated tokens on Qwen2.5-7B and Llama3.1-8B. HybridThinker consistently produces shorter outputs than Vanilla, H$_2$O, and LightThinker.}
    \label{fig:output_length}
\vspace{-5pt}
\end{figure}

Figure~\ref{fig:output_length} compares the average number of generated tokens across different methods. HybridThinker produces the shortest outputs on both Qwen2.5-7B and Llama3.1-8B, with 2,518 and 2,559 tokens respectively, which are lower than those of Vanilla, H$_2$O, and LightThinker. 
This indicates that HybridThinker alleviates redundant reasoning.
In contrast, H$_2$O produces the longest average output length, consistent with prior findings that KV cache pruning strategies at low budgets can paradoxically produce longer reasoning traces~\citep{liu2025holdthoughtassessingkv}, suggesting that KV cache pruning strategies lead to redundant reasoning.

\subsection{Case Study}
\label{sec:case_study_analysis}
\begin{figure}[h]
    \centering
    \includegraphics[width=1.0\linewidth]{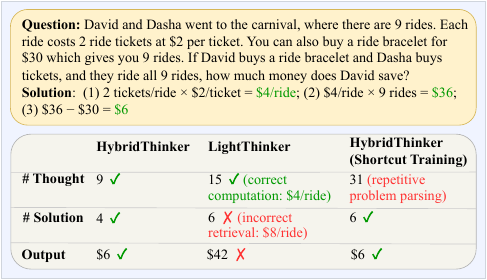}
    \caption{A case study comparing HybridThinker, LightThinker, and HybridThinker (Shortcut Training) on a mathematical problem. \textit{\# Thought} and \textit{\# Solution} denote the number of steps in the thinking and solution stages, respectively. Due to the length of the full model outputs, complete responses are provided in Appendix~\ref{sec:case_study_title}.}
    \label{fig:case_study}
\vspace{-5pt}
\end{figure}

Figure~\ref{fig:case_study} presents a case study on a mathematical problem. HybridThinker, through rigorous reasoning, completes the correct reasoning in just 13 steps. 
In contrast, LightThinker correctly computes the per-ride cost (\$4/ride) during the thinking stage, yet erroneously retrieves it as \$8/ride from its compressed memory during the solution stage, ultimately yielding an incorrect final answer of \$42.
Although the HybridThinker trained with Shortcut Attention also produces the correct answer, it requires as many as 37 steps, including 20 `\texttt{wait}' tokens, repeatedly parsing the problem statement rather than building on its compressed memory. This happens because Shortcut Attention training leaves the memory tokens' compression and retrieval insufficiently trained; at inference, the model therefore falls back to re-reading the question instead of trusting the compact representations to continue reasoning.
By contrast, the HybridThinker trained with Hybrid Attention needs to parse the problem only once, demonstrating that Hybrid Attention training effectively mitigates redundant reasoning.



\section{Conclusion}
We propose HybridThinker, which akin to CoT compression condenses thought steps into memory tokens but temporarily retains thought steps to supply fine-grained information together with memory tokens for inference. We discover that while such inference process is effective generally, training the model that aligns training and inference is suboptimal due to shortcut learning. To address this, we propose a hybrid training scheme mixing shortcut and bottleneck learning in attention. Experiments on 4 benchmarks show a 5.8 point average accuracy gain over existing CoT compression methods, matching the uncompressed baseline while reducing peak token usage by 61.7\% and inference time by 20.3\%. Compared to existing CoT compression methods, HybridThinker achieves higher average accuracy with similar inference time and slightly more ($\sim 10\%$) peak memory usage.

\section*{Limitations}
Temporarily retaining thought steps slightly increases peak KV cache usage compared to methods that discard steps immediately.
As shown in Figure~\ref{fig:hyperparameter_analysis}(a), this can be effectively mitigated by reducing the number of memory tokens $L$: with $L=3$, HybridThinker already achieves lower peak usage than LightThinker while maintaining a 2.5 point accuracy advantage.
Our approach also assumes explicit step delimiters; extending it to support free-form reasoning without clear boundaries is left for future work.


\bibliography{custom}

\clearpage

\appendix

\section{Experimental Details}

\subsection{Baseline Details}
\label{appendix:baseline_details}

During evaluation, we employ greedy decoding with a maximum output length of 10,240 tokens for all models. We compare HybridThinker against the following baselines:

\begin{itemize}[topsep=4pt, itemsep=4pt, parsep=2pt, leftmargin=*]
    \item \textbf{CoT}: A baseline that applies few-shot Chain-of-Thought (CoT) prompting to the Qwen2.5-7B~\cite{hui2024qwen2} and Llama-3.1-8B~\cite{dubey2024llama} models without additional training.

    \item \textbf{Distill-R1}~\cite{Guo_2025}: A reasoning model distilled from DeepSeek-R1's response data, built upon the Qwen and Llama architectures.
    
    \item \textbf{Vanilla}: A standard full-parameter instruction-tuned model. Operating without any compression or acceleration mechanisms, it serves as the empirical upper bound for reasoning accuracy.

    \item \textbf{H$_2$O}~\cite{zhang2023h2o}: A training-free KV cache eviction strategy that retains "Heavy Hitter" tokens and a local window. We apply H$_2$O to the Vanilla model using a sliding window of 1024 and a heavy-hitter budget of 512 tokens.

    \item \textbf{SepLLM}~\cite{chen2025sepllmacceleratelargelanguage}: A training-free framework that preserves KV caches for initial tokens, separators, and a local window. We configure SepLLM with an initial cache size of 384, a separator budget of 64, and a local window of 256, maintaining a total cache capacity of 1024.

    \item \textbf{LightThinker}~\cite{zhang2025lightthinker}: A post-training method that compresses each reasoning step into a fixed number of memory tokens. Upon reaching a step boundary, the model generates memory tokens to summarize the context, after which the original reasoning tokens are evicted from the KV cache. The training data and most hyperparameter settings follow those of HybridThinker, with the exception that LightThinker uses $L=9$, $w=1$, and $|\mathcal{I}|=0$, while all remaining hyperparameters are kept identical to HybridThinker.
\end{itemize}

\subsection{Training Details}
\label{appendix:train_details}

The Vanilla baseline, LightThinker and HybridThinker are initialized from DeepSeek-R1-Distill~\cite{Guo_2025} and trained on the Bespoke-Stratos-17k (BS17K) dataset for 5 epochs. Experiments are conducted on 8 H200 GPUs using DeepSpeed ZeRO-3 offload. We use a micro-batch size of 2 and 4 gradient accumulation steps, yielding a global batch size of 64. We employ a cosine learning rate schedule with a 0.05 warmup ratio. The peak learning rate for Vanilla is set to 1e-5, while for the other models, it is set to 2e-5. All the CoT token length (including memory tokens) has been set to 4096. For HybridThinker, we use 9 memory tokens, a retention horizon of $w=4$, and a shortcut set size of $|\mathcal{I}|=2$.

\subsection{Evaluation Details}

During inference, we evaluate all models using greedy decoding with a repetition penalty of 1.1. Prompt configurations for Table~\ref{table:exp_main} are as follows: Vanilla, H$_2$O, SepLLM, LightThinker, and HybridThinker share the same system prompt (Figure~\ref{fig:system_prompt_oth}) and task prompts (Figure~\ref{fig:task_prompt_oth}). Distill-R1 uses these task prompts but omits the system prompt. The CoT baseline uses a few-shot system prompt (Figure~\ref{fig:system_prompt_cot}) alongside task-specific prompts for different benchmarks (Figure~\ref{fig:task_prompt_cot}). For MMLU~\cite{hendrycks2021measuring} and GPQA~\cite{rein2023gpqa}, multiple-choice options are randomized to prevent positional bias.

\section{Case Study}
\label{sec:case_study_title}

\subsection{HybridThinker: Clear and Faithful Reasoning}
\label{sec:case_study}
As shown in Case~\ref{case:HybridThinker}, HybridThinker reasons in a clear, linear fashion. After framing the task as a cost comparison (Steps 1–2), it computes ticket cost at \$4 per ride and Dasha's total at \$36 (Steps 3–6), then directly concludes \$36 - \$30 = \$6 with a brief cross-verification (Steps 7–9). The solution phase faithfully mirrors this derivation (Steps 11–13). Brief self-corrections appear but are quickly resolved, leaving no redundant re-derivation. This coherent reasoning is enabled by HybridThinker's temporary retention of thought steps, which supplies subsequent steps with the fine-grained local details needed for accurate retrieval and computation.

\subsection{LightThinker: Retrieval Error from Compressed Memory}
As shown in Case~\ref{case:LightThinker}, LightThinker correctly identifies the per-ride cost as \$4 during the thinking stage (Steps 3--5), resolves the ambiguity of ride splitting (Steps 8--10), and confidently concludes \$36 - \$30 = \$6 with repeated verification (Steps 11--15). However, when transitioning to the solution stage, the model erroneously retrieves the per-ride cost as \$8 from its compressed memory (Step 17), computing 9 × \$8 = \$72 and ultimately outputting \$42 (Step 20), directly contradicting its own prior reasoning. This is consistent with the known limitation of existing CoT compression methods: immediately discarding thought steps after compression causes fine-grained details to be obscured or lost in the compact representations, making subsequent steps more error-prone.

\subsection{HybridThinker (shortcut): Redundant Reasoning Loop}
As shown in Case~\ref{case:HybridThinker(shortcut)}, this model arrives at the correct answer but at significant token cost. After correctly computing the \$6 saving in Steps 1–5, the model immediately derails into an extended loop (Steps 6–25) over the irrelevant question of how David and Dasha split the 9 rides—repeatedly re-reading the problem, self-contradicting, and failing to escape despite having already reached the correct conclusion. Throughout this loop, the token "wait" appears dozens of times in rapid succession as a reflexive filler that restarts rather than advances reasoning (highlighted in the original trace). The model only breaks free in Steps 26–30 by essentially ignoring the self-imposed ambiguity and re-confirming \$36 - \$30 = \$6, followed by a clean solution phase (Steps 32–35). This behavior is consistent with the shortcut learning problem described in our paper: when thought steps remain directly accessible during training, the model over-relies on plain-text thoughts rather than learning to compress and retrieve information through memory tokens, leaving the memory token mechanism undertrained and ultimately degrading reasoning coherence.

\begin{figure*}[htbp]
    \centering
    \begin{tcolorbox}[
        colback=yellow!8,      
        colframe=black!70,  
        arc=2mm,            
        boxrule=1.0pt,      
        left=4pt, right=4pt, top=4pt, bottom=4pt 
    ]
    \vspace{0.2cm}
    \textbf{System Prompt:} \\
    Below is a question. Please think through it step by step, and then provide the final answer. If options are provided, please select the correct one.
    \vspace{0.3cm}
    
    \#\# Output format: \\
    Use ``\textless THOUGHT\textgreater...\textless/THOUGHT\textgreater'' to outline your reasoning process, and enclose the final answer in `\textbackslash boxed\{\}'.
    \vspace{0.3cm}
    
    \#\# Example 1: \\
    Question: \\
    What is 2 + 3? \\
    Output: \\
    \textless THOUGHT\textgreater First, I recognize that this is a simple addition problem. Adding 2 and 3 together gives 5.\textless/THOUGHT\textgreater \\
    Therefore, the final answer is \textbackslash boxed\{5\}.
    \vspace{0.3cm}
    
    \#\# Example 2: \\
    Question: \\
    What is 2 + 3? \\
    A. 4 \\
    B. 5 \\
    C. 10 \\
    
    Output: \\
    \textless THOUGHT\textgreater First, I recognize that this is a simple addition problem. Adding 2 and 3 together gives 5.\textless/THOUGHT\textgreater \\
    Therefore, the final answer is \textbackslash boxed\{B\}.
    \vspace{0.2cm}
    \end{tcolorbox}
    \caption{System prompt for Qwen2.5-7B-Instruct and Llama3.1-8B-Instruct.}
    \label{fig:system_prompt_cot}
\end{figure*}

\begin{figure*}[htbp]
    \centering
    \begin{tcolorbox}[
        colback=yellow!8,      
        colframe=black!70,     
        arc=2mm,               
        boxrule=1.0pt,         
        left=4pt, right=4pt, top=4pt, bottom=4pt 
    ]
    \vspace{0.2cm}
    \textbf{GSM8K:} \\
    Return your final response within \textbackslash boxed\{\}. \textcolor{purple}{\{Question\}}
    \vspace{0.4cm}
    
    \textbf{MMLU:} \\
    Please select the option that best answers the question. Return your final response within \textbackslash boxed\{\}. Here are the Question: \\
    \textcolor{purple}{\{Question\}}
    \vspace{0.4cm}
    
    \textbf{GPQA:} \\
    Given a question, please select the option that best answers it. Return your final response within \textbackslash boxed\{\}. \textcolor{purple}{\{Question\}}
    \vspace{0.4cm}
    
    \textbf{BBH:} \\
    Return your final response within \textbackslash boxed\{\}. If options are provided, please select the correct one. \\
    \textcolor{purple}{\{Question\}}
    \vspace{0.2cm}
    \end{tcolorbox}
    \caption{Task prompt for Qwen2.5-7B-Instruct and Llama3.1-8B-Instruct.}
    \label{fig:task_prompt_cot}
\end{figure*}

\begin{figure*}[htbp]
    \centering
    \begin{tcolorbox}[
        colback=yellow!8,      
        colframe=black!70,     
        arc=2mm,               
        boxrule=1.0pt,         
        left=4pt, right=4pt, top=4pt, bottom=4pt 
    ]
    \vspace{0.2cm}
    \textbf{System Prompt:} \\
    Your role as an assistant involves thoroughly exploring questions through a systematic long thinking process before providing the final precise and accurate solutions. This requires engaging in a comprehensive cycle of analysis, summarizing, exploration, reassessment, reflection, backtracking, and iteration to develop well-considered thinking process. Please structure your response into two main sections: Thought and Solution. In the Thought section, detail your reasoning process using the specified format: \textless|begin\_of\_thought|\textgreater\ \{thought with steps separated with `\textbackslash n\textbackslash n'\} \textless|end\_of\_thought|\textgreater\ Each step should include detailed considerations such as analyzing questions, summarizing relevant findings, brainstorming new ideas, verifying the accuracy of the current steps, refining any errors, and revisiting previous steps. In the Solution section, based on various attempts, explorations, and reflections from the Thought section, systematically present the final solution that you deem correct. The solution should remain a logical, accurate, concise expression style and detail necessary steps needed to reach the conclusion, formatted as follows: \textless|begin\_of\_solution|\textgreater\ \{final formatted, precise, and clear solution\} \textless|end\_of\_solution|\textgreater\ Now, try to solve the following question through the above guidelines:
    \vspace{0.2cm}
    \end{tcolorbox}
    \caption{The shared system prompt applied to Vanilla, H$_2$O, SepLLM, LightThinker, and HybridThinker across the Qwen and Llama series.}
    \label{fig:system_prompt_oth}
\end{figure*}

\begin{figure*}[htbp]
    \centering
    \begin{tcolorbox}[
        colback=yellow!8,      
        colframe=black!70,     
        arc=2mm,               
        boxrule=1.0pt,         
        left=4pt, right=4pt, top=4pt, bottom=4pt 
    ]
    \vspace{0.2cm}
    \textbf{GSM8K/MMLU/GPQA/BBH:} \\
    Return your final response within \textbackslash boxed\{\}. \textcolor{purple}{\{Question\}}
    \vspace{0.2cm}
    \end{tcolorbox}
    \caption{The shared task prompt applied to Vanilla, H$_2$O, SepLLM, LightThinker, and HybridThinker across the Qwen and Llama series.}
    \label{fig:task_prompt_oth}
\end{figure*}

\onecolumn
\refstepcounter{casestudy}
\refstepcounter{casestudy}
\refstepcounter{casestudy}
\refstepcounter{casestudy}
\begin{tcolorbox}[
  breakable,
  title={B.4 Case: HybridThinker},
  colback=yellow!8,
  colframe=black!70,
  fonttitle=\bfseries,
  label={case:HybridThinker}
]
\textbf{Question:}\\
David and Dasha went to the carnival, where there are 9 rides.  Each ride costs 2 ride tickets at \$2 per ticket.  You can also buy a ride bracelet for \$30 which gives you 9 rides.  If David buys a ride bracelet and Dasha buys tickets, and they ride all 9 rides, how much money does David save?\\ \\
\textbf{Model Output:}\\
\texttt{<|begin\_of\_thought|>} \\
Okay, let's see. So David and Dasha went to the carnival, right? There are 9 rides. Each ride normally costs 2 ride tickets, and each ticket is \$2. But then there's this deal: a ride bracelet that gives you 9 rides for \$30. Hmm. So David buys the ride bracelet, and Dasha buys individual tickets. They both ride all 9 rides. The question is asking how much money David saves compared to if he had bought tickets instead.<|splitter|>\\
\\Wait, actually, David buys the bracelet, so he gets all 9 rides covered by that. Dasha buys tickets, so she has to pay for each ride separately. Since they're riding all 9 rides, we need to calculate how much each of them spent and then find the difference in what David would have spent on tickets versus the bracelet.<|splitter|>\\
\\First, let me break down the costs. Let's start with David. He buys the ride bracelet for \$30, which gives him 9 rides. So his total cost is \$30. That part seems straightforward.<|splitter|>\\
\\Now, Dasha buys tickets. Each ride costs 2 tickets, and each ticket is \$2. Wait, hold on. So each ride requires 2 tickets, and each ticket is \$2. So per ride, the cost is 2 * \$2 = \$4 per ride? Wait, no, maybe I misread that. Let me check again.<|splitter|>\\
\\The problem says: "Each ride costs 2 ride tickets at \$2 per ticket." So, each ride needs 2 tickets, and each ticket is \$2. Therefore, per ride, the cost is 2 tickets * \$2 per ticket = \$4 per ride. So for one ride, Dasha pays \$4. Since they take all 9 rides, but Dasha is buying individual tickets for each ride. Wait, but wait, the problem states that David buys a ride bracelet and Dasha buys tickets. The ride bracelet gives 9 rides for \$30. So Dasha is buying tickets for each ride, but how many tickets does she need?<|splitter|>\\
\\Each ride requires 2 tickets, so for 9 rides, Dasha needs 9 rides * 2 tickets per ride = 18 tickets. At \$2 per ticket, that's 18 * \$2 = \$36. So Dasha spends \$36 on tickets. Then David spends \$30 on the bracelet. Together, their total cost is \$30 + \$36 = \$66. But the question is asking how much David saves. So David's saving would be the difference between what he would have spent without the bracelet vs. what he actually spent. Wait, but David didn't buy tickets; he bought the bracelet. So perhaps we need to compare how much David saved by purchasing the bracelet instead of individual tickets for all 9 rides.<|splitter|>\\
\\Wait, but David only took all 9 rides once. If he had bought individual tickets, how much would he have paid? For 9 rides, each needing 2 tickets, so 18 tickets. 18 tickets * \$2 per ticket = \$36. But David bought the bracelet for \$30, which gives him 9 rides. So his saving is \$36 - \$30 = \$6. Therefore, David saved \$6 by buying the bracelet instead of individual tickets.<|splitter|>\\
\\But wait, let me confirm. The problem says "how much money does David save?" So since David chose to buy the bracelet instead of individual tickets, his saving is the amount he would have spent on tickets minus what he actually spent (the bracelet). So yes, \$36 - \$30 = \$6. Therefore, David saves \$6.<|splitter|>\\
\\Alternatively, another way to think about it: The bracelet is \$30 for 9 rides. If David were to buy individual tickets for those 9 rides, he would have to spend 9 rides * 2 tickets/ride * \$2/ticket = 9*2*2 = \$36. By buying the bracelet, he only pays \$30, so he saves \$6. Hence, the answer is \$6.<|splitter|>\\
\\I think that makes sense. Let me just make sure there's no other angle here. The problem mentions that they ride all 9 rides. So both David and Dasha are riding all 9. But David uses the bracelet, Dasha uses individual tickets. The question is specifically about how much David saves. So yes, comparing David's actual spending (\$30) to what he would have spent on tickets (\$36), the saving is \$6. So the answer should be \$6.<|splitter|>\\
\texttt{<|end\_of\_thought|>}\\
\\ \texttt{<|begin\_of\_solution|>}
\\David buys a ride bracelet for \$30, granting him access to all 9 rides. Dasha purchases individual ride tickets at \$2 per ticket, requiring ($9 \times 2 = 18$) tickets (($18 \times \$2 = \$36$)). <|splitter|>\\
\\To determine David's savings: 
\begin{itemize}
\item \textbf{Cost if David bought individual tickets}: ($18 \times \$2 = \$36$) 
\item \textbf{Actual cost}: \$30 (bracelet) 
\item \textbf{Savings}: (\$36 - \$30 = \$6)<|splitter|>\\
\end{itemize}
Thus, David saves $\boxed{6}$ dollars.<|splitter|>\\
\texttt{<|end\_of\_solution|>}\\
\\ \texttt{<|answer|>}
\\$6$\\
\texttt{<|answer|>}
\end{tcolorbox}

\onecolumn
\refstepcounter{casestudy}
\begin{tcolorbox}[
  breakable,
  title={B.5 Case: LightThinker},
  colback=yellow!8,
  colframe=black!70,
  fonttitle=\bfseries,
  label={case:LightThinker}
]
\textbf{Question:}\\
David and Dasha went to the carnival, where there are 9 rides.  Each ride costs 2 ride tickets at \$2 per ticket.  You can also buy a ride bracelet for \$30 which gives you 9 rides.  If David buys a ride bracelet and Dasha buys tickets, and they ride all 9 rides, how much money does David save?\\ \\
\textbf{Model Output:}\\
\texttt{<|begin\_of\_thought|>} \\
Okay, let's see. So David and Dasha went to the carnival, right? There are 9 rides. Each ride normally costs 2 ride tickets, and each ticket is \$2. But then there's this deal: a ride bracelet that gives you 9 rides for \$30. Now, David buys the bracelet, and Dasha buys tickets. They ride all 9 rides. We need to figure out how much money David saves compared to if he had bought tickets instead.<|splitter|>\\
\\Hmm, first I need to understand what exactly is being asked here. The question is about how much David saves by buying the ride bracelet instead of individual tickets. So we need to compare the cost of the bracelet to the cost if David had bought tickets for all 9 rides.
<|splitter|>\\
\\Let me break it down. Let's start with David. He buys the ride bracelet. The bracelet costs \$30 and gives him 9 rides. Since they both end up riding all 9 rides, David doesn't need to buy individual tickets. So his total cost is just \$30.
<|splitter|>\\
\\Now, Dasha is buying tickets. Each ride costs 2 tickets, but wait, the tickets are sold individually at \$2 each. Wait, hold on. Let me check the problem again. It says each ride costs 2 ride tickets at \$2 per ticket. So each ride requires 2 tickets, and each ticket is \$2. Wait, that seems a bit confusing. Wait, maybe I misread that. Let me parse that sentence again.
<|splitter|>\\
\\Each ride costs 2 ride tickets at \$2 per ticket. Hmm. So each ride requires 2 tickets, and each ticket costs \$2. Therefore, per ride, the cost would be 2 tickets * \$2 per ticket = \$4 per ride. Is that right? So each ride is \$4?
<|splitter|>\\
\\Wait, but then if they take the bracelet, it's \$30 for 9 rides. So let's confirm the costs.
<|splitter|>\\
\\If they don't buy the bracelet, each ride costs 2 tickets at \$2 each, so \$4 per ride. For 9 rides, that would be 9 * \$4 = \$36. But Dasha is only buying tickets, so she might not need all 9 rides' tickets? Wait, no. The problem states they ride all 9 rides. So both David and Dasha each ride 4.5 rides? Wait, no, that can't be. Wait, there are 9 rides total. So David and Dasha split the rides? Or do they go on the rides together?
<|splitter|>\\
\\Wait, the problem says "they ride all 9 rides". So between David and Dasha, they use all 9 rides. But how are the rides divided? Are they splitting the rides, or is one of them riding all 9? The problem doesn't specify. Hmm. That could affect the calculation. Wait, but the question is about how much David saves by buying the bracelet. So perhaps the key is that David buys the bracelet, which gives him 9 rides, and Dasha buys tickets for her own rides. But since they have 9 rides total, maybe David takes some number of rides and Dasha takes the rest? But the problem doesn't specify how many rides each takes. Wait, maybe the problem assumes that David uses the bracelet, which gives him 9 rides, and Dasha uses tickets for her own rides, but since they have 9 rides total, maybe Dasha is riding some number of rides. Wait, this is unclear.
<|splitter|>\\
\\Wait, perhaps the problem is that David buys the bracelet (which allows him to get 9 rides) and Dasha buys individual tickets for her rides. Then, since there are 9 rides total, David takes some number of rides and Dasha takes the remaining. But the problem states they ride all 9 rides. So maybe David takes x rides and Dasha takes (9 - x) rides. However, the problem asks how much David saves by buying the bracelet. So David's saving would be the difference between the cost of the tickets he would have bought for his rides versus the bracelet. But since David is part of a group (probably going together), maybe they split the rides. Wait, but the problem doesn't specify how the rides are split between David and Dasha. Hmm.
<|splitter|>\\
\\Wait, maybe the problem is intended to assume that David takes the bracelet, which gives him 9 rides, and Dasha buys tickets for however many rides she needs. But since they have to ride all 9 rides, perhaps David is taking the bracelet, which covers all 9 rides, and Dasha doesn't need to buy any tickets. But then why does the problem mention Dasha buying tickets? Maybe there's a misunderstanding here.
<|splitter|>\\
\\Alternatively, perhaps the problem is structured such that David buys the bracelet, which allows him to ride all 9 rides, and Dasha buys individual tickets for all 9 rides. But that would mean Dasha is buying 9 * 2 tickets (since each ride costs 2 tickets). Wait, but each ticket is \$2, so 2 tickets per ride would be 2 * \$2 = \$4 per ride. So 9 rides would cost Dasha 9 * \$4 = \$36. Meanwhile, David paid \$30 for the bracelet. So the total amount they spent would be \$30 + \$36 = \$66. But the question is asking how much David saves. Wait, David saved money by buying the bracelet instead of individual tickets. How much would he have spent on tickets? If he were to buy tickets for 9 rides, each ride costing 2 tickets at \$2 each, so per ride it's \$4. So 9 rides would be 9 * \$4 = \$36. But David didn't buy tickets; he bought the bracelet for \$30. So his saving is \$36 - \$30 = \$6. Therefore, David saves \$6.
<|splitter|>\\
\\But wait, the problem says "how much money does David save?" So David bought the bracelet, which costs \$30, whereas if he had bought tickets for all 9 rides, he would have had to spend 9 rides * \tcbox[colframe=teal, colback=yellow!8, boxrule=0.8pt, on line, left=1pt, right=1pt, top=0pt, bottom=0pt]{\$4 per ride} = \$36. Therefore, by buying the bracelet, he saved \$36 - \$30 = \$6. So David's saving is \$6.
<|splitter|>\\
\\But let me verify once more. The key point is that David chose to buy the bracelet instead of individual tickets. Since the bracelet gives all 9 rides for \$30, while buying individual tickets would cost 9 rides * 2 tickets per ride * \$2 per ticket = \$36. Therefore, David's saving is \$36 - \$30 = \$6. So yes, David saves \$6 by purchasing the bracelet.
<|splitter|>\\
\\I think that makes sense. The key is that David chooses the bracelet, which covers all the rides he and Dasha plan to take, thereby avoiding the cost of the individual tickets. Since the bracelet is a fixed price regardless of the number of rides (as long as it's 9), he saves the difference between the total cost of individual tickets and the bracelet price.
<|splitter|>\\
\\So the answer should be \$6, which is the amount David saved by purchasing the bracelet instead of individual tickets for all 9 rides.
<|splitter|>\\
\texttt{<|end\_of\_thought|>}\\
\\ \texttt{<|begin\_of\_solution|>}
\\David purchases a ride bracelet for \$30, which entitles him to all 9 rides. If he were to buy individual tickets instead, each ride costs 2 tickets at \$2 per ticket, \tcbox[colframe=red!70, colback=yellow!8, boxrule=0.8pt, on line, left=1pt, right=1pt, top=0pt, bottom=0pt]{totaling \$8 per ride} (\$16 for 2 rides). For 9 rides, this would amount to ($9 \times \$8 = \$72$).<|splitter|>\\
\\By purchasing the bracelet, David avoids the cost of these individual tickets. The savings are calculated as:[\$72 - \$30 = \$42]
<|splitter|>\\
\\However, upon re-evaluating the problem statement, it becomes clear that David and Dasha share the rides. Since the bracelet covers all 9 rides, David’s saving specifically relates to the tickets he would have used. If David rode all 9 rides individually, his total cost would have been \$72. By purchasing the bracelet, he only pays \$30, resulting in a saving of:[\$72 - \$30 = \$42]<|splitter|>\\
\\Thus, David saves $\boxed{42}$ dollars.
<|splitter|>\\
\texttt{<|end\_of\_solution|>}\\
\\ \texttt{<|answer|>}
\\$6$\\
\texttt{<|answer|>}
\end{tcolorbox}

\onecolumn
\refstepcounter{casestudy}
\begin{tcolorbox}[
  breakable,
  title={B.6 Case: HybridThinker (Shortcut Training)},
  colback=yellow!8,
  colframe=black!70,
  fonttitle=\bfseries,
  label={case:HybridThinker(shortcut)}
]
\textbf{Question:}\\
David and Dasha went to the carnival, where there are 9 rides.  Each ride costs 2 ride tickets at \$2 per ticket.  You can also buy a ride bracelet for \$30 which gives you 9 rides.  If David buys a ride bracelet and Dasha buys tickets, and they ride all 9 rides, how much money does David save?\\ \\
\textbf{Model Output:}\\
\texttt{<|begin\_of\_thought|>} \\
Okay, let's see. So David and Dasha went to the carnival, right? There are 9 rides. Each ride normally costs 2 ride tickets, and each ticket is \$2. But they can also buy a ride bracelet for \$30 that gives you 9 rides. Now, David buys the ride bracelet, and Dasha buys tickets. They ride all 9 rides. The question is asking how much money David saves compared to if he had bought tickets instead.<|splitter|>\\
\\First, I need to figure out how much David spent and how much he would have spent if he didn't buy the bracelet. Then, the difference between those two amounts is his savings.<|splitter|>\\
\\\tcbox[colframe=red!70, colback=yellow!8, boxrule=0.8pt, on line, left=1pt, right=1pt, top=0pt, bottom=0pt]{wait}, actually, the \tcbox[colframe=red!70, colback=yellow!8, boxrule=0.8pt, on line, left=1pt, right=1pt, top=0pt, bottom=0pt]{problem says} David buys a ride bracelet and Dasha buys tickets. So David's cost is the price of the bracelet, which is \$30. But if he had bought individual tickets instead, how much would he have paid?<|splitter|>\\
\\\tcbox[colframe=red!70, colback=yellow!8, boxrule=0.8pt, on line, left=1pt, right=1pt, top=0pt, bottom=0pt]{wait}, but each ride costs 2 tickets at \$2 per ticket. So each ride is 2 tickets * \$2 = \$4 per ride. Since there are 9 rides total, but David and Dasha split the rides somehow? \tcbox[colframe=red!70, colback=yellow!8, boxrule=0.8pt, on line, left=1pt, right=1pt, top=0pt, bottom=0pt]{wait}, no, \tcbox[colframe=red!70, colback=yellow!8, boxrule=0.8pt, on line, left=1pt, right=1pt, top=0pt, bottom=0pt]{the problem states} they ride all 9 rides. So both of them together ride all 9 rides. Hmm, but how are the tickets split between David and Dasha?<|splitter|>\\
\\But \tcbox[colframe=red!70, colback=yellow!8, boxrule=0.8pt, on line, left=1pt, right=1pt, top=0pt, bottom=0pt]{wait}, maybe not. Let me read again. "David buys a ride bracelet and Dasha buys tickets." So David's expense is the bracelet, which covers all 9 rides. Dasha's expense is tickets for some rides. But the problem doesn't specify how many rides each of them takes. \tcbox[colframe=red!70, colback=yellow!8, boxrule=0.8pt, on line, left=1pt, right=1pt, top=0pt, bottom=0pt]{wait}, but it says they ride all 9 rides. So combined, they use all 9 rides. But since David has the bracelet, which gives him 9 rides, maybe Dasha doesn't need to pay for any rides? That can't be right because \tcbox[colframe=red!70, colback=yellow!8, boxrule=0.8pt, on line, left=1pt, right=1pt, top=0pt, bottom=0pt]{the problem says} Dasha buys tickets. Maybe they each buy their own tickets? \tcbox[colframe=red!70, colback=yellow!8, boxrule=0.8pt, on line, left=1pt, right=1pt, top=0pt, bottom=0pt]{wait}, no. \tcbox[colframe=red!70, colback=yellow!8, boxrule=0.8pt, on line, left=1pt, right=1pt, top=0pt, bottom=0pt]{Let me parse the problem again}.<|splitter|>\\
\\\tcbox[colframe=red!70, colback=yellow!8, boxrule=0.8pt, on line, left=1pt, right=1pt, top=0pt, bottom=0pt]{wait}, \tcbox[colframe=red!70, colback=yellow!8, boxrule=0.8pt, on line, left=1pt, right=1pt, top=0pt, bottom=0pt]{the problem says}: "If David buys a ride bracelet and Dasha buys tickets, and they ride all 9 rides, how much money does David save?"<|splitter|>\\
\\ "Each ride costs 2 ride tickets at \$2 per ticket." So each ride requires 2 tickets, each ticket costing \$2, so per ride, it's \$4. Alternatively, you can buy a bracelet for \$30 that gives you 9 rides. So the bracelet is \$30 for 9 rides, whereas buying individual tickets would be 9 rides * 2 tickets each * \$2 per ticket. Let's calculate that. 9 rides * 2 tickets = 18 tickets. Each ticket is \$2, so 18 * \$2 = \$36. So the total cost for 9 rides via tickets is \$36, while the bracelet is \$30. So buying the bracelet is cheaper by \$6.\tcbox[colframe=red!70, colback=yellow!8, boxrule=0.8pt, on line, left=1pt, right=1pt, top=0pt, bottom=0pt]{wait}, the key information:<|splitter|>\\
\\But in this scenario, David buys the bracelet, and Dasha buys tickets. How do they split the cost? Or is the question about how much David saves by buying the bracelet instead of individual tickets?<|splitter|>\\
\\But \tcbox[colframe=red!70, colback=yellow!8, boxrule=0.8pt, on line, left=1pt, right=1pt, top=0pt, bottom=0pt]{the problem states} that David buys the bracelet and Dasha buys tickets. Then they ride all 9 rides. The question is how much David saves. So perhaps we need to compare how much David would have spent if he bought tickets versus the bracelet.<|splitter|>\\
\\\tcbox[colframe=red!70, colback=yellow!8, boxrule=0.8pt, on line, left=1pt, right=1pt, top=0pt, bottom=0pt]{wait}, but David is buying the bracelet, so he spends \$30. If he were to buy individual tickets for the rides he took, how much would that be? \tcbox[colframe=red!70, colback=yellow!8, boxrule=0.8pt, on line, left=1pt, right=1pt, top=0pt, bottom=0pt]{wait}, but how many rides did David take? The problem doesn't specify. It just says they ride all 9 rides. So maybe they split the rides somehow? But the problem doesn't mention splitting the rides; it just says they ride all 9. So perhaps both David and Dasha take some rides, but the total is 9. However, the problem statement isn't clear on how the rides are divided between them. 	\tcbox[colframe=red!70, colback=yellow!8, boxrule=0.8pt, on line, left=1pt, right=1pt, top=0pt, bottom=0pt]{wait}, maybe \tcbox[colframe=red!70, colback=yellow!8, boxrule=0.8pt, on line, left=1pt, right=1pt, top=0pt, bottom=0pt]{the problem assumes} that David uses the bracelet, which gives him 9 rides, and Dasha buys tickets to cover her own rides. But since the total number of rides is 9, maybe David takes all 9 rides? But then why would Dasha buy tickets? This part is confusing.<|splitter|>\\
\\\tcbox[colframe=red!70, colback=yellow!8, boxrule=0.8pt, on line, left=1pt, right=1pt, top=0pt, bottom=0pt]{wait}, \tcbox[colframe=red!70, colback=yellow!8, boxrule=0.8pt, on line, left=1pt, right=1pt, top=0pt, bottom=0pt]{perhaps the problem is} that David buys the bracelet (which allows him to ride 9 times) and Dasha buys individual tickets for herself. But since they only have 9 rides total, David takes all 9 rides? But then Dasha wouldn't need any tickets. Alternatively, maybe they each take some rides, totaling 9. But the problem doesn't specify how the rides are distributed between them. This is ambiguous.<|splitter|>\\
\\\tcbox[colframe=red!70, colback=yellow!8, boxrule=0.8pt, on line, left=1pt, right=1pt, top=0pt, bottom=0pt]{wait}, maybe the problem is simpler. \tcbox[colframe=red!70, colback=yellow!8, boxrule=0.8pt, on line, left=1pt, right=1pt, top=0pt, bottom=0pt]{Let's re-read it}:<|splitter|>\\
\\\tcbox[colframe=red!70, colback=yellow!8, boxrule=0.8pt, on line, left=1pt, right=1pt, top=0pt, bottom=0pt]{wait}, \tcbox[colframe=red!70, colback=yellow!8, boxrule=0.8pt, on line, left=1pt, right=1pt, top=0pt, bottom=0pt]{the problem says}: "If David buys a ride bracelet and Dasha buys tickets, and they ride all 9 rides, how much money does David save?" So David buys the bracelet, Dasha buys tickets. Together, they use the rides. The total rides used are 9. We need to find how much David saved by buying the bracelet instead of tickets.<|splitter|>\\
\\So David's saving is the difference between what he would have spent on tickets versus the bracelet. But how much would David have spent on tickets? If he was buying tickets for the rides he took. But how many rides did he take? The problem doesn't specify. Similarly, Dasha took some rides. But without knowing how many rides each took, how can we compute the saving?<|splitter|>\\
\\But maybe the problem assumes that David took all 9 rides, but that contradicts because Dasha is buying tickets. 	\tcbox[colframe=red!70, colback=yellow!8, boxrule=0.8pt, on line, left=1pt, right=1pt, top=0pt, bottom=0pt]{wait}, no. If David buys the bracelet, he gets 9 rides, but Dasha is buying tickets. So Dasha might be taking some rides, but she needs tickets for each. But the total rides are 9. So maybe all 9 rides are taken by David and Dasha together. But how is the cost calculated?<|splitter|>\\
\\\tcbox[colframe=red!70, colback=yellow!8, boxrule=0.8pt, on line, left=1pt, right=1pt, top=0pt, bottom=0pt]{wait}, \tcbox[colframe=red!70, colback=yellow!8, boxrule=0.8pt, on line, left=1pt, right=1pt, top=0pt, bottom=0pt]{perhaps the problem is} intended to be that David buys the bracelet, which covers all 9 rides, so he pays \$30. Dasha buys tickets for the remaining rides? But there are no remaining rides. All 9 rides are covered by the bracelet. Therefore, David saved the amount he would have spent on 9 rides minus the bracelet cost.<|splitter|>\\
\\\tcbox[colframe=red!70, colback=yellow!8, boxrule=0.8pt, on line, left=1pt, right=1pt, top=0pt, bottom=0pt]{wait}, but David didn't need tickets for any rides because he has the bracelet. So his saving is the cost of the bracelet compared to the cost of the tickets he would have bought for the same number of rides. But how many rides is that?<|splitter|>\\
\\\tcbox[colframe=red!70, colback=yellow!8, boxrule=0.8pt, on line, left=1pt, right=1pt, top=0pt, bottom=0pt]{wait}, \tcbox[colframe=red!70, colback=yellow!8, boxrule=0.8pt, on line, left=1pt, right=1pt, top=0pt, bottom=0pt]{maybe the problem is that} David could have bought 9/2 rides worth of tickets, but that doesn't make sense. Alternatively, maybe the problem is that David and Dasha split the cost of the rides. But \tcbox[colframe=red!70, colback=yellow!8, boxrule=0.8pt, on line, left=1pt, right=1pt, top=0pt, bottom=0pt]{the problem states} that David buys the bracelet and Dasha buys tickets. So David's cost is \$30, Dasha's cost is tickets for some rides. But how much is that?<|splitter|>\\
\\Alternatively, \tcbox[colframe=red!70, colback=yellow!8, boxrule=0.8pt, on line, left=1pt, right=1pt, top=0pt, bottom=0pt]{maybe the problem is} that David wants to ride all 9 rides, but instead of buying the bracelet himself, he's buying tickets, but Dasha is also buying tickets. But \tcbox[colframe=red!70, colback=yellow!8, boxrule=0.8pt, on line, left=1pt, right=1pt, top=0pt, bottom=0pt]{the problem says} David buys a ride bracelet and Dasha buys tickets. So David's saving is the difference between the cost of 9 rides via tickets vs. the bracelet.<|splitter|>\\
\\\tcbox[colframe=red!70, colback=yellow!8, boxrule=0.8pt, on line, left=1pt, right=1pt, top=0pt, bottom=0pt]{wait}, perhaps the answer is simply the difference between the cost of 9 rides via tickets (which is 9 rides * 2 tickets * \$2 per ticket) versus the bracelet (\$30). So 9*2*2=36. Bracelet is \$30. So David saves \$6. Therefore, the answer is \$6.<|splitter|>\\
Therefore, David's saving is \$36 - \$30 = \$6.<|splitter|>\\
But let me check again. If David buys the bracelet, he pays \$30 for 9 rides. If he bought tickets instead, he would need 9 rides, each requiring 2 tickets, so 18 tickets. At \$2 per ticket, 18*2 = \$36. So he saves \$6. Therefore, David saves \$6 by buying the bracelet instead of tickets.<|splitter|>\\
\\But \tcbox[colframe=red!70, colback=yellow!8, boxrule=0.8pt, on line, left=1pt, right=1pt, top=0pt, bottom=0pt]{the problem says} David buys a bracelet and Dasha buys tickets. So Dasha is buying tickets for some rides. But the total rides are 9. So if David is already getting all 9 rides via the bracelet, then Dasha doesn't need to buy any tickets. But \tcbox[colframe=red!70, colback=yellow!8, boxrule=0.8pt, on line, left=1pt, right=1pt, top=0pt, bottom=0pt]{the problem says} Dasha buys tickets. Maybe there's a misinterpretation here.<|splitter|>\\
\\Alternatively, maybe David and Dasha each want to ride 4.5 times, which isn't possible, so one rides 4 and the other 5. But since you can't split rides, maybe the problem expects us to assume that David rides x times and Dasha rides (9 - x) times. Then, David's cost would be the cost of the bracelet plus the cost of x rides via tickets, but no, \tcbox[colframe=red!70, colback=yellow!8, boxrule=0.8pt, on line, left=1pt, right=1pt, top=0pt, bottom=0pt]{the problem says} David buys a bracelet and Dasha buys tickets. So David's total cost is \$30, Dasha's cost is 2*(number of tickets she buys). The total rides are 9, so the sum of David's rides and Dasha's rides is 9. But David's rides are covered by the bracelet, so David doesn't need to buy any tickets. Therefore, David's cost is only \$30, and Dasha's cost is 2*tickets. But how many tickets does Dasha buy? If all 9 rides are taken, then David's rides via bracelet is 9, so Dasha doesn't need any tickets. \tcbox[colframe=red!70, colback=yellow!8, boxrule=0.8pt, on line, left=1pt, right=1pt, top=0pt, bottom=0pt]{But the problem says} Dasha buys tickets. This is confusing.<|splitter|>\\
\\This suggests that maybe the problem is structured differently. Perhaps David decides to go on some rides using the bracelet, and Dasha goes on the others using tickets. For example, David buys the bracelet and chooses to ride 5 times, thereby paying \$30, and Dasha buys tickets for the remaining 4 rides. Then, Dasha would pay 4*2*2 = \$16. But the problem doesn't specify how many rides each took. It just says they rode all 9 rides. So \tcbox[colframe=red!70, colback=yellow!8, boxrule=0.8pt, on line, left=1pt, right=1pt, top=0pt, bottom=0pt]{maybe the problem is} that David could have bought tickets for all 9 rides, which would cost \$36, but instead he bought a bracelet for \$30, saving \$6. Therefore, regardless of how the rides are split between David and Dasha, David's saving is \$6. Because even though Dasha is buying tickets, David's saving is independent of how the rides are divided; he just saved \$6 by choosing the bracelet over tickets.<|splitter|>\\
\\\tcbox[colframe=red!70, colback=yellow!8, boxrule=0.8pt, on line, left=1pt, right=1pt, top=0pt, bottom=0pt]{wait}, but if David buys the bracelet, he gets all 9 rides, so he doesn't need to spend anything else. Dasha buys tickets for some rides. But \tcbox[colframe=red!70, colback=yellow!8, boxrule=0.8pt, on line, left=1pt, right=1pt, top=0pt, bottom=0pt]{the problem states} that they ride all 9 rides. So perhaps the total cost for David is \$30, and the total cost for Dasha is for the remaining rides. But since \tcbox[colframe=red!70, colback=yellow!8, boxrule=0.8pt, on line, left=1pt, right=1pt, top=0pt, bottom=0pt]{the problem asks} how much David saves, it's only comparing David's cost with what he would have spent if he didn't buy the bracelet. But if he bought tickets instead, he would have to pay for the number of rides he actually took. But since the problem doesn't specify how many rides David took, only that they rode all 9, maybe the answer is that David saved the difference between the bracelet and the total tickets he would have bought, assuming he took all 9. But that seems like an assumption.<|splitter|>\\
\\Alternatively, maybe \tcbox[colframe=red!70, colback=yellow!8, boxrule=0.8pt, on line, left=1pt, right=1pt, top=0pt, bottom=0pt]{the problem is} intended to be straightforward: David saves the cost of 9 rides via tickets instead of the bracelet. So 9 rides at 2 tickets each is 18 tickets, which is \$36. The bracelet is \$30, so he saves \$6. Therefore, the answer is \$6.<|splitter|>\\
\\Yes, that must be it. \tcbox[colframe=red!70, colback=yellow!8, boxrule=0.8pt, on line, left=1pt, right=1pt, top=0pt, bottom=0pt]{The problem is} likely testing the understanding that the bracelet is a flat rate for all rides, whereas tickets are per ride. So by buying the bracelet, David saves the cost of 9 rides' worth of tickets. Therefore, the saving is \$36 - \$30 = \$6.<|splitter|>\\
\\I think that's the correct approach. So David saves $\boxed{6}$ dollars.<|splitter|>
\texttt{<|end\_of\_thought|>}\\
\\ \texttt{<|begin\_of\_solution|>}
\\To determine how much David saves by purchasing a ride bracelet instead of individual tickets, we first calculate the total cost for 9 rides using tickets. Each ride requires 2 tickets at \$2 per ticket, so the cost per ride is ($2 \times 2 = \$4$). For 9 rides, the total cost would be ($9 \times 4 = \$36$).<|splitter|>\\
\\David opts for a ride bracelet that costs \$30, which provides all 9 rides. The difference between the cost of individual tickets and the bracelet is:<|splitter|>\\
\\By purchasing the bracelet, David avoids the full cost of 9 tickets. Therefore, his savings are calculated as:<|splitter|>\\
\\\text{Savings} = \text{Cost of 9 tickets} - \text{Cost of bracelet} = \$36 - \$30 = \$6<|splitter|>\\
\\Thus, David saves $\boxed{6}$ dollars.<|splitter|>\\
\texttt{<|end\_of\_solution|>}\\
\\ \texttt{<|answer|>}
\\$6$\\
\texttt{<|answer|>}
\end{tcolorbox}

\end{document}